\documentclass[10pt,twocolumn,letterpaper]{article}

\usepackage{cvpr}              % To produce the CAMERA-READY version
\definecolor{cvprblue}{rgb}{0.21,0.49,0.74}
\usepackage[pagebackref,breaklinks,colorlinks,allcolors=cvprblue]{hyperref}
\usepackage{amsmath}
\usepackage{amssymb}
\usepackage{times}  % DO NOT CHANGE THIS
\usepackage{helvet}  % DO NOT CHANGE THIS
\usepackage{courier}  % DO NOT CHANGE THIS
\usepackage[hyphens]{url}  % DO NOT CHANGE THIS
\usepackage{graphicx} % DO NOT CHANGE THIS
\usepackage{natbib}  % DO NOT CHANGE THIS AND DO NOT ADD ANY OPTIONS TO IT
\usepackage{caption} % DO NOT CHANGE THIS AND DO NOT ADD ANY OPTIONS TO IT
\usepackage[table]{xcolor}
\usepackage{multirow}
\usepackage{booktabs}
\usepackage{algorithm}
\usepackage{algorithmic}

\usepackage{newfloat}
\usepackage{listings}
\usepackage{bibentry}

\newcommand{\figref}[1]{Fig.~\ref{#1}}
\newcommand{\tabref}[1]{Table~\ref{#1}}

\def\paperID{22572} % *** Enter the Paper ID here
\def\confName{CVPR}
\def\confYear{2026}

\title{High-Precision Dichotomous Image Segmentation\\ via Depth Integrity-Prior and Fine-Grained Patch Strategy}

\author{Xianjie Liu$^1$ ~~~ Keren Fu$^{1,2,}$\thanks{Corresponding author: Keren Fu (fkrsuper@scu.edu.cn).} ~~~ Qijun Zhao$^{1,2}$\\
$^1$College of Computer Science, Sichuan University, Chengdu, China\\
$^2$National Key Lab of Fundamental Science on Synthetic Vision, Sichuan University}

\begin{document}
\maketitle

\begin{abstract}
High-precision dichotomous image segmentation (DIS) is a task of extracting fine-grained objects from high-resolution images.
Existing methods trade efficiency for accuracy: non-diffusion methods are fast but suffer from weak semantics and unstable spatial priors, causing false detections; diffusion-based methods offer high accuracy via strong generative priors but are computationally expensive.
In depth maps, a complete object appears as a low variance region with a smooth interior and sharp boundaries, whereas the background exhibits a chaotic, high variance pattern due to disconnected surfaces at varying depths. We refer to this as the depth integrity-prior.
Inspired by this, and noting that DIS currently lacks depth maps, we leverage pseudo-depth information from monocular depth estimation models to obtain essential semantic understanding, thereby rapidly revealing spatial differences across target objects and the background.
To exploit this prior, we propose the Prior-guided Depth Fusion Network (PDFNet), which fuses RGB and pseudo-depth features for depth-aware structure perception. We further introduce a novel depth integrity-prior loss to enforce depth consistency in segmentation and a fine-grained enhancement module with adaptive patch selection to sharpen boundaries.
Notably, PDFNet with DAM-v2 achieves SOTA \textbf{($F^{max}_\beta$ 0.915 on DIS-VD and 0.915 on DIS-TE)} using less than half the params of diffusion-based methods.
Our code is available at \href{https://tennine2077.github.io/PDFNet.github.io/}{https://tennine2077.github.io/PDFNet.github.io/}.
\end{abstract}

%%%%%%%%%%%%%%%%%%%%% Figure2:inte-prior %%%%%%%%%%%%%%%%%%%%%
\begin{figure}
  \centering
  \includegraphics[width=0.46\textwidth]{fig/depth-v.png}
  \centering
  \includegraphics[width=0.46\textwidth]{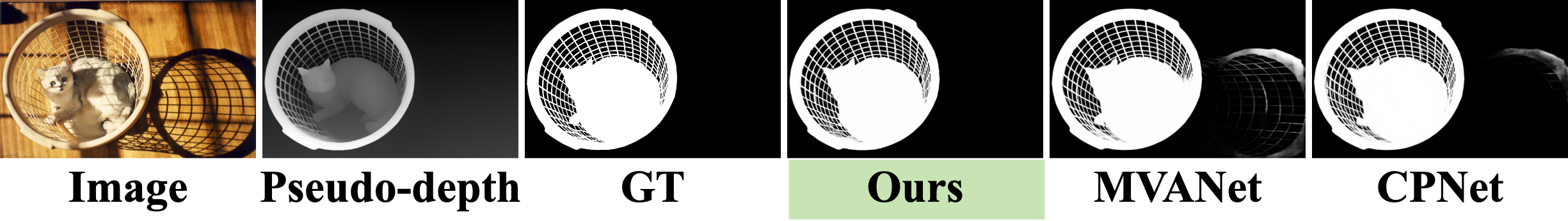}
  \caption{
We generate pseudo depth for the DIS-TR dataset and compute depth variance within ground truth (GT) regions, background regions, and the full map. GT regions exhibit significantly \textbf{lower} depth variance. Visually, our method outperforms both CPNet (RGB-D model) and MVANet (DIS model).
 }
 % \vspace{-5pt}
  \label{fig:inte-prior}
\end{figure}
%%%%%%%%%%%%%%%%%%%%% Figure %%%%%%%%%%%%%%%%%%%%

\section{Introduction}
High-precision dichotomous image segmentation (DIS) \cite{IS-net} is a critical computer vision task aiming to precisely delineate foreground objects from high-resolution (HR) images at the pixel level.
Achieving such meticulous segmentation is increasingly vital for numerous real-world applications requiring high-fidelity human-computer interaction, including image editing \cite{dis-imageediting,IC-Light} and augmented reality \cite{AR1,AR2}.
While advances in digital imaging technology have made HR image acquisition easily accessible, converting these rich visual details into accurate masks remains extremely challenging, especially under complex conditions.

Research on DIS has predominantly followed two major technical routes: the non-diffusion paradigm and the diffusion paradigm. Non-diffusion methods (such as Convolutional Neural Networks \cite{IS-net,UDUN,FP-DIS} and Transformer architectures \cite{InSPyReNet,BiRefNet,MVANet}) usually have advantages of lightweight  ($>$10M and $<$300M) and faster inference speeds (FPS$>$3) compared with diffusion methods. However, they encounter a fundamental bottleneck on high-resolution images: 
when the receptive field is enlarged to capture global structures, the capability of modeling fine details is weakened. On the contrary, when the receptive field is concentrated to preserve local details, the modeling of global structures is less satisfactory \cite{MVANet}.
Ultimately, constrained by this bottleneck, the model’s weak semantics and lack of robust spatial priors cause segmentation results to frequently exhibit false or missed detections.
Diffusion methods \cite{genpercept,diffdis} introduce ultra-large-scale pre-trained diffusion models as backbone networks and adopt post-training to utilize prior information from billions of image data. Such methods significantly improve segmentation consistency in complex scenarios. But the cost is a huge number of params ($>$865M) and extremely slow inference speeds (FPS$<$1). This characteristic makes them infeasible in practical application scenarios.

%%%%%%%%%%%%%%%%%%%% Figure2:overall %%%%%%%%%%%%%%%%%%%%%
\begin{figure}
  \centering
  \includegraphics[width=0.48\textwidth]{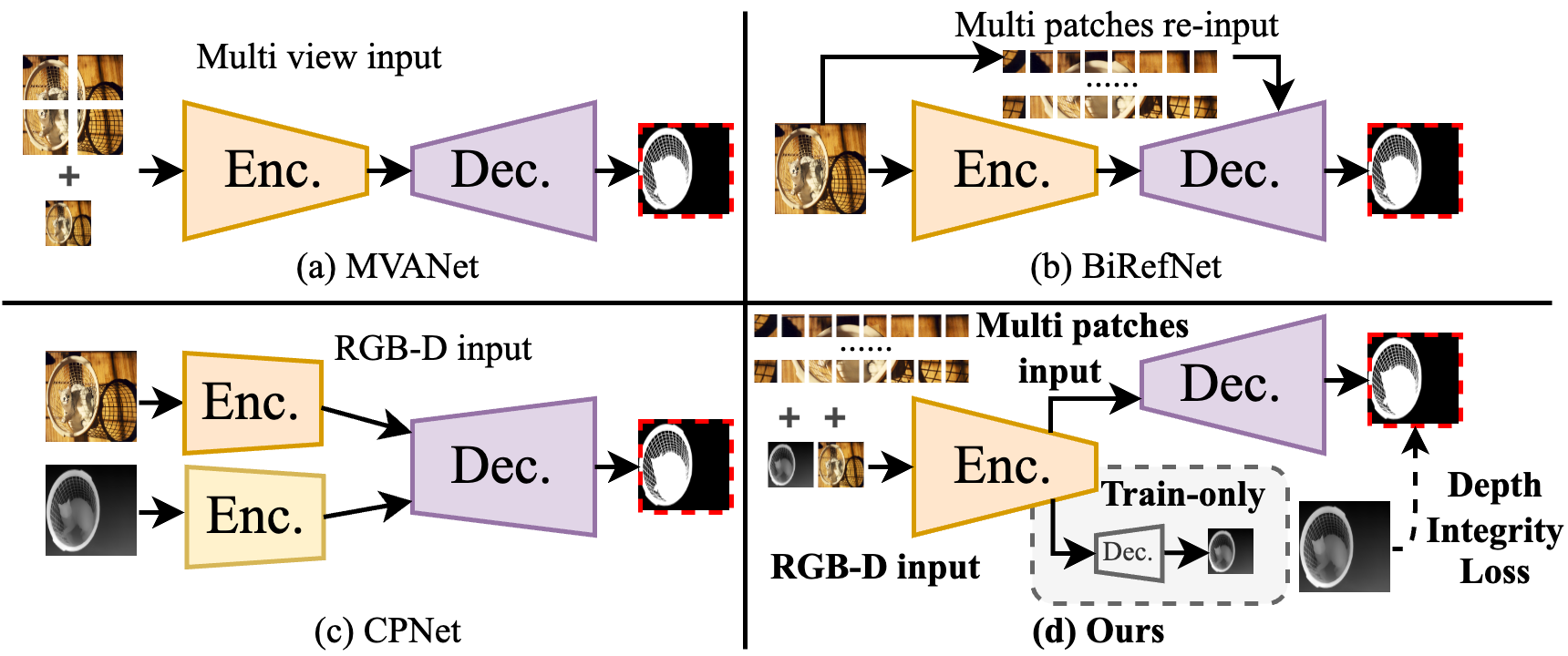}
  \caption{Comparison between our PDFNet and other methods. (a) MVANet \cite{MVANet}: adopts 4 patches and image for multi-view input. (b) BiRefNet \cite{BiRefNet}: input the image into the encoder and re-inputs the image patches into the decoder. (c) CPNet \cite{CPNet}: employs dual-modal joint learning. (d) Ours: input the image, depth map, and multi patches, add using the depth integrity-prior loss and depth-refinement. Enc. = Encoder, Dec. = Decoder.
 }
 % \vspace{-5pt}
  \label{fig:framwork}
\end{figure}
%%%%%%%%%%%%%%%%%%%% Figure2:overall %%%%%%%%%%%%%%%%%%%%%

To overcome this trade off, we need a task-adaptive prior that meets three key criteria: \textbf{easy accessibility} (can be derived at low cost from existing reliable models), \textbf{high performance} (with a small number of params and fast inference speed), and \textbf{strong guidance} (capable of clearly distinguishing between objects and the background).

We observe that in depth maps, a complete object appears as a low variance region with smooth interior and sharp boundaries, while the background exhibits high variance, chaotic patterns due to disconnected surfaces at varying depths (Fig. \ref{fig:inte-prior}). We refer to this as the depth integrity-prior.
Motivated by this, we extract key semantic cues from pseudo-depth maps generated by Depth Anything Model v2 (DAM-v2) \cite{depth_anything_v2} to effectively reveal spatial distinctions between foreground objects and background, providing \textbf{strong guidance} for DIS. The pseudo-depth maps offer \textbf{easy accessibility} obtainable directly from off-the-shelf DAM-v2 and \textbf{high performance}, with DAM-v2-Base running at $>$ 10 FPS.
To better exploit this prior, we propose the Prior-guided Depth Fusion Network (PDFNet), illustrated in Figure \figref{fig:framwork}d. To our knowledge, this is the first work in DIS to incorporate depth as a modality.
PDFNet is tailored for high-resolution, fine-grained object segmentation by fusing the depth integrity-prior. It models multimodal interaction via cross-modal attention to enable depth-guided structural awareness. We further introduce a novel depth integrity-prior loss that enforces consistency in the predicted mask by constraining the mean and edges of the pseudo-depth within foreground regions. Additionally, we extend MVANet \cite{MVANet} from 2×2 to 8×8 patches and integrate a fine-grained perception enhancement module with adaptive patch selection to refine boundary-sensitive details.
On the DIS-5K dataset, PDFNet achieves \textbf{SOTA} performance, matching the diffusion-based DiffDIS \cite{diffdis} while using less than 50\% of its params and outperforms all non-diffusion methods.

Our main contributions can be summarized as follows:
\begin{itemize}
    \item We introduce depth as a new modality to DIS and propose the depth integrity-prior, which provides strong spatial guidance. Given the prevalence of multimodal learning, this offers a reference for future DIS research.
    \item We introduce a novel depth integrity-prior loss to enhance depth-wise consistency of the segmentation output.
    \item We design a fine-grained perception module with adaptive patch selection. By increasing patch density (to 8x8), it enhances boundary-sensitive detail refinement while effectively suppressing features from non-target areas.
    \item We demonstrate that PDFNet, a non-diffusion paradigm, achieves \textbf{SOTA} performance compared to diffusion methods with a fraction ($<$50\%) of the params.
\end{itemize}

\section{Related Works}
\subsection{Dichotomous Image Segmentation}
High-Precision Dichotomous Image Segmentation (DIS) aims to delineate intricate objects in complex scenes, a formidable challenge.
Initial methods set foundational benchmarks \cite{IS-net}. Subsequent non-diffusion approaches sought improvements through multi-scale refinement \cite{UDUN,InSPyReNet}, frequency prior \cite{FP-DIS}, and multi-view analysis \cite{BiRefNet,MVANet}. More recently, diffusion models have emerged, leveraging powerful generative priors from large-scale datasets to enhance segmentation quality \cite{genpercept,diffdis}.
The DIS task is also closely related to High-Resolution Salient Object Detection (HRSOD) \cite{HRSOD, pyramid}. While both fields leverage similar strategies to handle high-resolution inputs, their objectives differ. HRSOD, as seen in methods like PGNet \cite{pyramid}, focuses on visually dominant objects by targeting saliency. In contrast, DIS is a more generalized task aiming for a complete and precise separation of the primary foreground subject(s). This places a greater emphasis on capturing intricate structures and achieving high boundary fidelity, regardless of the object's visual saliency.
However, a common thread across these paradigms is a critical trade-off. Conventional methods struggle to balance global and local cues, causing false or missed detections \cite{MVANet}. Conversely, diffusion models, while accurate, incur prohibitive computational costs that limit their practical use. We resolve this trade-off with our proposed PDFNet. Our framework introduces a novel \textit{depth integrity-prior} to enforce structural consistency. This prior, combined with adaptive local patch fusion, enables high-fidelity segmentation with remarkable efficiency.

%%%%%%%%%%%%%%%%%%%%% Figure2:overall %%%%%%%%%%%%%%%%%%%%%
\begin{figure*}
  \centering
  \includegraphics[width=0.98\textwidth]{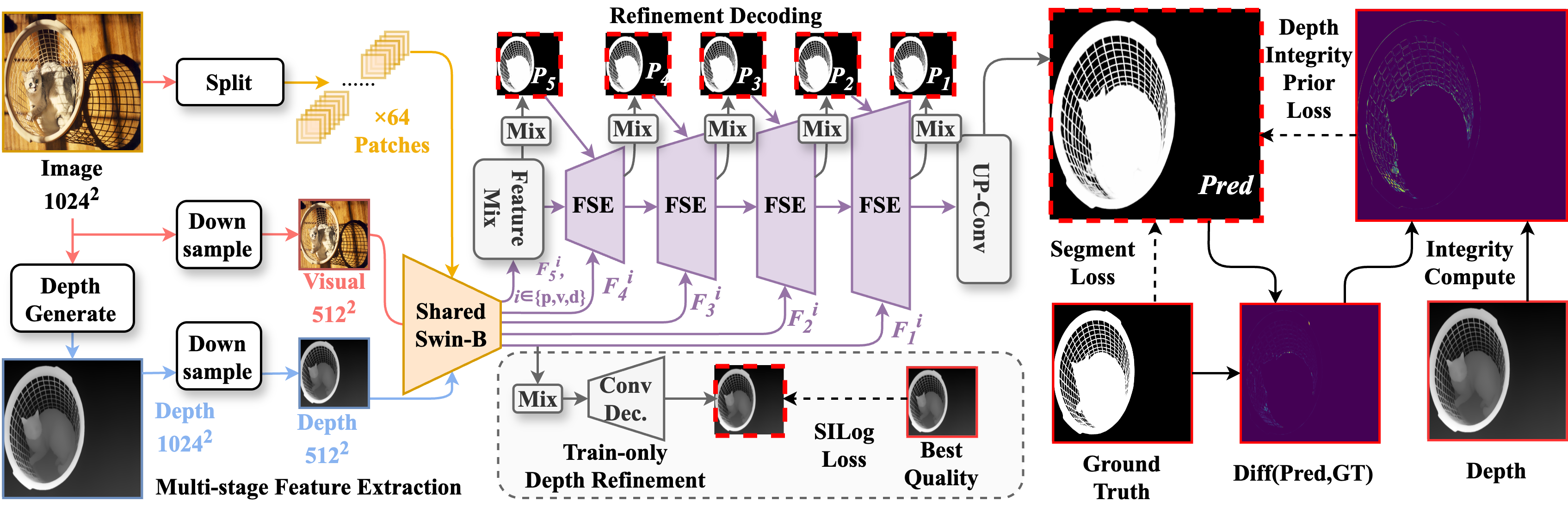}
  
  \caption{
 Overall pipeline of the proposed PDFNet.
 }
 
  \label{fig:overall}
\end{figure*}

%%%%%%%%%%%%%%%%%%%%% Figure %%%%%%%%%%%%%%%%%%%%%

\subsection{Monocular Depth Estimation}  
Monocular depth estimation, a foundational task in computer vision, infers scene depth from a single RGB image. Early deep learning works pioneered end-to-end solutions using self-supervised disparity signals \cite{monodepth17}. A paradigm shift occurred with large-scale models like the Depth Anything Model (DAM), which uses Transformers and massive pre-training to boost depth accuracy and global coherence \cite{depth_anything_v1}. Its successor, DAM-v2, addresses the real-data annotation bottleneck via knowledge distillation\cite{depth_anything_v2}. With SOTA performance (e.g., 97.1\% on DA-2K \cite{depth_anything_v2}), DAM-v2 has become a standard for generating reliable pseudo-depth.

\subsection{Real/Pseudo Depth Maps for Dense Prediction}
Depth-aware methods enhance dense prediction using real or pseudo-depth cues. Real-depth methods fuse sensor-derived geometry with RGB data \cite{magnet, CPNet}. In contrast, pseudo-depth methods generate depth from RGB inputs, proving efficacy in instance segmentation \cite{boosting}, medical imaging \cite{polyp-DAM}, and beyond \cite{ssfSOD, popnet, impact, IOCUS, Segment_any_motion}. For DIS, as no real-depth   exists before, we use pseudo-depth to introduce our \textit{depth integrity-prior}. This prior is operationalized within a framework featuring specialized fusion modules and a dedicated loss to enhance structural coherence.

\section{Methodology}
In this section, we will present the approach, including the overall and specific components as shown in \figref{fig:overall}. 

\subsection{Overall Architecture}
\subsubsection{Depth Generation}
We choose DAM-v2 \cite{depth_anything_v2} to be the pseudo-depth generator. It maps input images $I \in \mathbb{R}^{B \times 3 \times H \times W}$ to a normalized depth map $D \in \mathbb{R}^{B \times 1 \times H \times W}$ within the range [0, 1].

\subsubsection{Multi-stage Feature Extraction}
Inspired by shared encoder frameworks \cite{MVANet, depthpro}, our model employs a cross-modal architecture to facilitate feature interaction. Given high-resolution RGB images $I \in \mathbb{R}^{B \times 3 \times H \times W}$ and corresponding depth maps $D \in \mathbb{R}^{B \times 1 \times H \times W}$, a main encoder extracts multi-scale visual and depth features, $\{F^v_i\}_{i=1}^4$ and $\{F^d_i\}_{i=1}^4$, respectively. To capture fine-grained details, a parallel branch partitions the input images into 64 patches, reorganized as a batch of size $\mathbb{R}^{64 \times B \times 3 \times \frac{H}{8} \times \frac{W}{8}}$. A dedicated patch encoder then processes these patches to yield features $\{F^{pj}_i\}_{j=0}^{63}$, which are subsequently reassembled into a high-resolution feature sequence $\{F^p_i\}_{i=1}^4$. Finally, a series of cross-scale $3\times3$ convolutions fuse these multi-level feature streams, producing the final representations $\{F^v_5, F^d_5, F^p_5\}$. In this design, $\{F^v_i, F^d_i\}$ provide global spatial context, while $\{F^p_i\}$ is responsible for high-fidelity detail representation. This multi-branch architecture is designed to enhance both modality complementarity and feature consistency.

\subsubsection{Refinement Decoding}
Diverging from the classic U-Net architecture, our decoder integrates a Feature Selection and Extraction (FSE) module at each stage. As illustrated in \figref{fig:overall}, these FSE modules are designed to dynamically enhance salient features. This process is conditioned on an analysis of boundary and integrity cues derived from the previous stage's predictions. Concurrently, the modules employ a cross-attention mechanism to progressively fuse multi-modal information. Furthermore, shallow features from the encoder are systematically integrated into the decoder's upsampling path to enrich contextual details and refine spatial accuracy.

\subsubsection{Depth Refinement}
To regularize the feature learning process, we introduce a depth refinement task implemented via a dedicated decoder. This strategy serves a dual purpose: it guides the shared encoder to learn representations beneficial for both segmentation and depth estimation, and it compels the model to extract fine-grained details from the RGB image through the pseudo-depth reconstruction objective. The depth refinement decoder itself has a simple architecture; each stage is composed of two stacked 3$\times$3 convolutional blocks with SiLU activations \cite{SiLU} and RMSNorm \cite{RMSNorm}, followed by a final 3$\times$3 convolutional layer to produce the depth prediction.

\subsubsection{Deep Supervision and Multi-feature Merging}
Following established practice \cite{IS-net,MVANet,BiRefNet}, we apply deep supervision at multiple stages. The final prediction $P$ is progressively upsampled and fused with shallow encoder features \cite{MVANet,swinir}, mirrored in the depth decoder.

%-------------------------------------------------------------------------
\subsection{Feature Selection and Extraction (FSE)}
\begin{figure}
  \centering
  \includegraphics[width=0.46\textwidth]{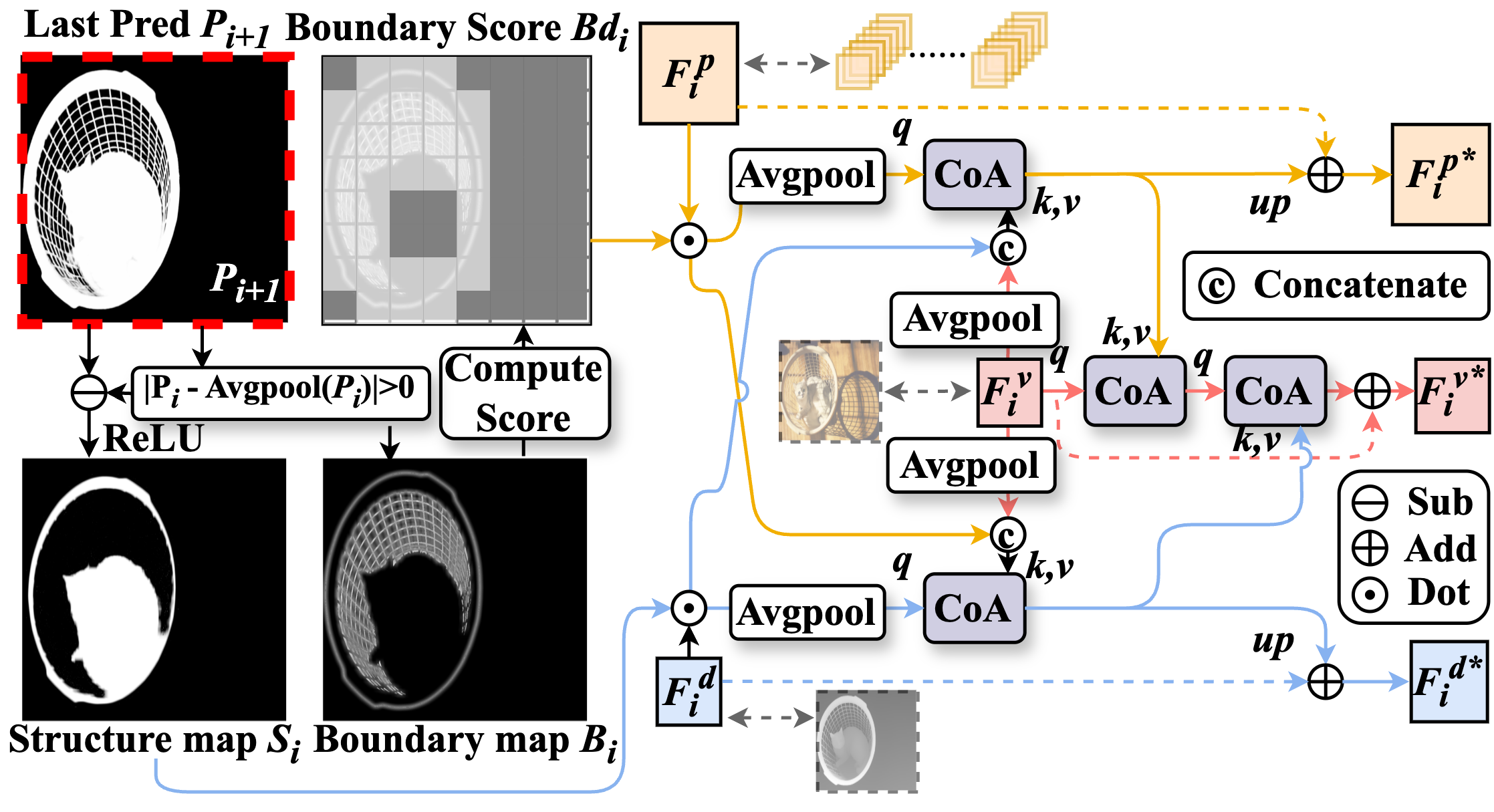}
  
  \caption{
Feature selection and extraction (FSE) module.
 }
 % \vspace{-5pt}
  \label{fig:components}
\end{figure}

The patch-based encoder improves detail extraction by limiting the receptive field, but loses contextual connections. To address this, we design the FSE module as shown in \figref{fig:components}, which dynamically enhances patches based on previous predictions to focus on target boundaries.

The FSE applies boundary-integrity separation on the previous prediction $P_{i+1} \in \mathbb{R}^{B \times 1 \times H \times W}$: an average-pooling operation is performed to obtain $P_{p_{i+1}}$, and the absolute difference yields a boundary map $B_i$, enhancing edge gradients. These operations are expressed as follows:
\begin{align}
P_{p_{i+1}} &=\text{AvgPool}(P_{i+1}),\\
{B_{i}}_{(x,y)} &= 1,\text{if }  |{P_{i+1}}_{(x,y)}-{P_{p_{i+1}}}_{(x,y)}| > \tau,else~0,
\end{align}
where the kernel size of the AvgPool is $(H/8, W/8)$ to adapt to inputs of different resolutions, and $\tau$ is a small constant set to 0.1. 
The original prediction result $P_{i+1}$ is subtracted from the binary boundary response $B_i$ and then processed through ReLU to obtain the target integrity map $S_i$ after boundary suppression, which focuses on the feature representation of the continuous regions inside the target.
\begin{equation}
S_i = \text{ReLU}(P_{i+1}-B_i).
\end{equation}
The strategy employs a difference amplification mechanism to enhance boundary sensitivity and maintains the consistency of the internal structure through nonlinear suppression. The boundary response map $B_i$ is divided into 64 patches, corresponding to the patch-based input. After binarization selection, the boundary response score $Bd_i$ for each patch is obtained. A weighting mechanism is used to assign corresponding weights to each patch, selectively enhancing features in patches that contain target boundaries. Given a binary boundary map $ B_i \in \{0,1\}^{H \times W} $, it is divided into $ N = 64 $ non-overlapping patches:
\begin{align}
    B_i^n &= \text{Patch}(B_i, n), \quad n = 0,1,2,...,63, \\
    Bd_i^n &= 1,~\text{if } \exists (x,y) \text{ in } B_i^n ,B_i^n(x,y) > 0,else~0,
\end{align}
where $ B_i^n \in \{0,1\}^{\frac{H}{8} \times \frac{W}{8}} $ denotes the $ n $-th patch.

%%%%%%%%%%%%%%%%%%%%% Figure3:components %%%%%%%%%%%%%%%%%%%%%

%%%%%%%%%%%%%%%%%%%%% Figure %%%%%%%%%%%%%%%%%%%%%

Secondly, to achieve effective multi-modal fusion, we propose the Cross-modal Attention (CoA) module. Based on cross-attention, CoA dynamically enhances complementary information between modalities through QKV interactions and Q/K projections. It includes: (1) computing attention between pixel sequences, (2) integrating original queries with RMSNorm-normalized attention outputs, and (3) applying a SwiGLU-based \cite{SwiGLu} FFN with residual connections to generate final features $F$. 
For memory efficiency, we compress original features $F^v_i$, $F^d_i$, $F^p_i$ using stage-specific average pooling to obtain $FP^v_i$, $FP^d_i$, $FP^p_i$ \cite{MVANet,p2t,asymmetric}. These are tokenized and concatenated into cross-modal embeddings $FP^{vd}_i$ (visual-depth) and $FP^{vp}_i$ (visual-patches), enabling continuous inter-modal interaction. Before CoA computation, image features are serialized for attention optimization.
Then, cross-modal feature associations are established through CoA, and the structural constraints of depth features and local detail features are dynamically integrated into the global context through $F^v_i$:
\begin{align}
FN^{p*}_i&=\text{CoA}(FP^p_i\odot (1+Bd_i),FP^{vd}_i),\\
FN^{d*}_i&=\text{CoA}(FP^d_i\odot (1+S_i),FP^{vp}_i),\\
FN^{v1*}_i&=\text{CoA}(F^{v*}_i,FN^{p*}_i),\\
FN^{v2*}_i&=\text{CoA}(FN^{v1*}_i,FN^{d*}_i).
\end{align}
Finally, the visual, patch and depth features are updated:
\begin{align}
F^{v*}_i &= F^v_i+FN^{v2*}_i,\\
F^{p*}_i &= F^p_i+\text{up}(FN^{p*}_i),\\
F^{d*}_i &= F^d_i+\text{up}(FN^{d*}_i).
\end{align}

\subsection{Depth Integrity-Prior Loss}
We exploit a key observation that depth values within a correctly segmented object region exhibit high internal consistency. We formalize this property as the \textbf{depth integrity-prior}, which reflects the structural coherence inherent in real-world objects. To enforce this prior during training, we introduce a loss function designed to penalize deviations from two core principles: internal depth stability and boundary-aligned depth continuity.

The first component, which we term the \textbf{depth stability constraint}, is motivated by the statistical distribution of depth values within the target mask. This constraint is engineered to mitigate two common error types: (1) false positives ($\mathcal{FP}$), where pixels with depths that deviate significantly from the object's mean are incorrectly included; and (2) false negatives ($\mathcal{FN}$), where pixels with depths consistent with the object's mean are erroneously omitted. To this end, our loss formulation adaptively weights the penalty for each pixel based on its depth deviation from the object's mean. A false positive with a large depth deviation from the mean incurs a high penalty. Conversely, a false negative with a small depth deviation is also heavily penalized, encouraging its inclusion.
Specifically, we first compute the mean depth $\mu$ of the ground-truth mask region $M$:
\begin{equation}
\mu={\sum(D\odot M)}/{\sum M}.
\end{equation}
The loss then selectively weights the standard cross-entropy term based on the squared depth difference, penalizing $\mathcal{FP}$ and $\mathcal{FN}$ regions differently:
\begin{equation}
l_v = \mathbb{E}[-\log\mathcal{P}_y\odot(\textit{diff}\odot (\mathcal{FP}-\mathcal{FN}) + \mathcal{FN})],
\end{equation}
where $D$ is the depth map, $P$ is the predicted probability map, and $\mathcal{P}_y$ represents the pixel-wise prediction correctness. The terms $\textit{diff}$, $\mathcal{FP}$, and $\mathcal{FN}$ are formulated as:
\begin{align}
\mathcal{P}_y &= P\odot M+(1-P)\odot (1-M),\\
    \textit{diff} &= (D-\mu)^2,\\
    \mathcal{FP} &= (1-\mathcal{P}_y) \odot P,\\
    \mathcal{FN} &= (1-\mathcal{P}_y) \odot M.
\end{align}
Since the $diff\in[0,1]$, the $l_v$ is non-negative.
The second component is the \textbf{depth continuity constraint}. This loss is motivated by the strong correlation between object boundaries and sharp discontinuities in depth maps. It enforces this relationship by up-weighting segmentation errors that occur at locations with high depth gradients. The corresponding loss term is designed to reduce spatial inconsistencies between the predicted mask and depth gradients:
\begin{equation}
l_{g}=\mathbb{E}[-\log\mathcal{P}_y\odot (|G_x|+|G_y|)],
\end{equation}
where $G_x$ and $G_y$ are the horizontal and vertical depth gradients, respectively, computed via the Sobel operator.
Together, these two constraints compel the model to leverage depth cues to learn a more structurally coherent representation, improving its ability to distinguish object interiors from boundaries. The final \textbf{depth integrity-prior loss} $l_{inte}$ is formulated as the average of these two components:
\begin{equation}
l_{inte} = ({l_{v} + l_{g}})/{2}.
\end{equation}

\subsection{Loss Function}
We supervise the output of each layer in the decoder as well as the final prediction result. In addition, we also supervise the result of each layer in the depth refinement decoder. The supervision for the output of each layer is denoted as $l^i_f$, and the supervision for the final result is denoted as $l_f$.
For segmentation supervision, following standard practices in segmentation tasks, we adopt a combination of weighted Binary Cross-Entropy \cite{wbce-wiou} ($l_{wBCE}$), weighted Intersection over Union \cite{wbce-wiou} ($l_{wIoU}$), and SSIM loss \cite{ssim} ($l_{SSIM}$) , following the practice of most segmentation tasks \cite{IS-net,MVANet,BiRefNet}, and our proposed depth integrity-prior loss ($l_{inte}$):
\begin{equation}
\begin{split}
 l = l_{wBCE} + l_{wIoU} + {l_{SSIM}}/{2} + l_{inte}.
\end{split}
\end{equation}
For the supervision of depth refinement, we use the Scale Invariant Logarithmic error \cite{SILogloss} (SILog) loss, which is adopted in most tasks.
Finally, our overall loss can be written in the following form:
\begin{equation}
\begin{split}
 L = l_f + \lambda_1 \cdot \sum^5_{i=1} l_f^i + \lambda_2 \cdot ( l_{SILog} + \lambda_1 \cdot \sum^5_{i=1} l_{SILog}^i),
\end{split}
\end{equation}
where $\lambda_1$ and $\lambda_2$ are set to 0.5 and 0.1, respectively.

%-----------------------------------comparewithsota--------------------------------------

\begin{table*}
\caption{Comparisons of PDFNet with IS-Net, FP-DIS, UDUN, InSPyReNet, BiRefNet, MVANet ,GenPercept, DiffDIS, MAGNet, CPNet and MVANet$^*$. The best is highlighted in \textbf{bold}, and the second is \underline {underlined} without diffusion methods because of the much larger params. Also note that the scores of DiffDIS \cite{diffdis} are somewhat special compared to other methods since it employs a ``pre-metric binarization'' step, which converts the resulting maps into binary maps, and such a step will decrease $F^{max}_{\beta}$ and $S_{\alpha}$ scores while raising other metrics. Since neither MVANet nor BiRefNet \cite{BiRefNet} uses pre-metric binarization, we choose to follow them and discard binarization.}
\label{tab:comparewithsota}
\centering
\small
\renewcommand{\arraystretch}{1}
\renewcommand{\tabcolsep}{0.8mm}
\begin{tabular}{ccc|ccccc|ccccc|ccccc}
 \toprule
 \multirow{2}{*}{\textbf{Methods}}& \multirow{2}{*}{\textbf{Modality}} & \multirow{2}{*}{\textbf{Params}} & \multicolumn{5}{c|}{\textbf{DIS-VD (470)}} & \multicolumn{5}{c|}{\textbf{DIS-TE1 (500)}} & \multicolumn{5}{c}{\textbf{DIS-TE2 (500)}} \\
 ~ & ~ & ~&$F^{max}_\beta$&$F^w_\beta$&$E_{\phi}^{m}$&$S_{\alpha}$&$M$& $F^{max}_\beta$&$F^w_\beta$&$E_{\phi}^{m}$&$S_{\alpha}$&$M$& $F^{max}_\beta$&$F^w_\beta$&$E_{\phi}^{m}$&$S_{\alpha}$&$M$\\
\hline
IS-Net$_{22}$& RGB     & 44M    & .791          & .717          & .856          & .813          & .074          & .740          & .662          & .820          & .787          & .074          & .799          & .728          & .858          & .823          & .070          \\
UDUN$_{23}$&    RGB    & 25M    & .823          & .763          & .891          & .843          & .062          & .784          & .720          & .864          & .817          & .059          & .829          & .768          & .886          & .843          & .058          \\
InSPyReNet$_{22}$& RGB & 88M   & .889          & .834          & .914          & .900          & .042          & .845          & .788          & .894          & .873          & .043          & .894          & .846          & .916          & .905          & .036          \\
BiRefNet$_{24}$&  RGB  & 215M    & .897          & .854          & .931          & .898          & .038          & .866          & .819          &  {.911}    &  {.885}    &  {.037}    & .906          & .857          & .930          & .900          & .036          \\
MVANet$_{24}$&   RGB   & 93M     & .904    & .856    & .938    &  {.905}    & .036    &  {.873}    &  {.823}    &  {.911}    & .879          &  {.037}    & .916    &  {.874}    &  {.944}    &  {.915}    &  {.030}    \\
\hline
\rowcolor[HTML]{D9D9D9} 
GenPercept$_{25}$& RGB & 865M+84M & .877          & .859          & .941          & .887          & .035          & .850          & .827          & .919          & .878          & .036          & .880          & .859          & .938          & .892          & .034          \\
\rowcolor[HTML]{D9D9D9} 
DiffDIS$_{25}$&  RGB   & 865M+84M & .908          & .888          & .948          & .904          & .029          & .883          & .862          & .933          & .891          & .030          & .917          & .895          & .951          & .913          & .026          \\
 \hline
MAGNet$_{24}$& RGB-D& 16M+335M  & .867 & .820 & .917 & .879 & .045 & .838 & .790 & .899 & .862 & .044 & .876 & .833 & .923 & .886 & .041 \\
CPNet$_{24}$& RGB-D   & 216M+335M & .892 & .855 & .933 & .900 &  {.034} & .862 & .819 & .910 & .880 &  {.037} & .892 & .855 & .929 & .899 & .036 \\

MVANet$^*$$_{24}$&   RGB-D   & 93M+335M     &  {.900}    & .856    &  {.931}    &  .906    & .033    & .881    & .837    & .922    & .894          & \underline {.031}    &  {.908}    &  {.869}    &  {.937}    &  {.913}    &  {.032}    \\

\hline
\textbf{PDFNet-S} &RGB-D&94M+24M & .909          & {.868}    & {.942}    & \underline{.913}    & \underline{.030} & .887          & .842          & \underline{.926}    & .896          & \underline{.031} & .915          & .877          & .944          & .918          & .030          \\
\textbf{PDFNet-B} &RGB-D &94M+97M& \underline{.912}    & \underline{.873} & \underline{.944} & \textbf{.916} & \underline{.030}       & \underline{.888}    & \underline{.844}    & \underline{.926}    & \underline{.898}    & \underline{.031} & \underline{.919}    & \underline{.883}    & \underline{.946}    & \underline{.922}    & \underline{.029}    \\
\textbf{PDFNet-L} & RGB-D & 94M+335M & \textbf{.915} & \textbf{.875} & \textbf{.945} & \textbf{.916} & \textbf{.029} & \textbf{.891} & \textbf{.848} & \textbf{.928} & \textbf{.899} & \textbf{.030} & \textbf{.920} & \textbf{.886} & \textbf{.948} & \textbf{.924} & \textbf{.027} \\

 % \hline
 % \toprule
 \bottomrule
 % \hline
 \multirow{2}{*}{\textbf{Methods}}& \multirow{2}{*}{\textbf{Modality}} & \multirow{2}{*}{\textbf{Params}} & \multicolumn{5}{c|}{\textbf{DIS-TE3 (500)}} & \multicolumn{5}{c|}{\textbf{DIS-TE4 (500)}} & \multicolumn{5}{c}{\textbf{DIS-TE (ALL) (2,000)}} \\
 % \hline
 ~ &  ~ & ~&$F^{max}_\beta$&$F^w_\beta$&$E_{\phi}^{m}$&$S_{\alpha}$&$M$& $F^{max}_\beta$&$F^w_\beta$&$E_{\phi}^{m}$&$S_{\alpha}$&$M$& $F^{max}_\beta$&$F^w_\beta$&$E_{\phi}^{m}$&$S_{\alpha}$&$M$\\
 
\hline

IS-Net$_{22}$&   RGB   & 44M    & .830          & .758          & .883          & .836          & .064          & .827          & .753          & .870          & .830          & .072          & .799          & .725          & .858          & .819          & .070          \\
UDUN$_{23}$&    RGB    & 25M    & .865          & .809          & .917          & .865          & .050          & .846          & .792          & .901          & .849          & .059          & .831          & .772          & .892          & .844          & .057          \\
InSPyReNet$_{22}$& RGB & 88M   & .919          & .871          & .940          & .918          & .034          & .905          & .848          & .936          &  { .905}          & .042          & .891          & .838          & .922          & .900          & .039          \\
BiRefNet$_{24}$&  RGB  & 215M    & .920          &  {.893}    & \underline {.955}    & .919          & \underline {.028}    &  {.906}    &  {.864}    & .939          & .900          &  {.039}    & .900          & .858          & .934          & .901          &  {.035}    \\
MVANet$_{24}$&   RGB   & 93M     &  {.929}    & .890          & .954          &  {.920}    & .031          & \textbf{.912}          & .857          & \textbf{.944}    & .903    & .041          &  {.908}    &  {.861}    &  {.938}    &  {.904}    &  {.035}    \\
\hline
\rowcolor[HTML]{D9D9D9} 
GenPercept$_{25}$& RGB & 865M+84M & .898          & .879          & .954          & .896          & .032          & .874          & .858          & .947          & .874          & .041          & .875          & .856          & .939          & .885          & .036          \\
\rowcolor[HTML]{D9D9D9}
DiffDIS$_{25}$&  RGB  & 865M+84M & .934          & .916          & .964          & .919          & .025          & .909          & .893          & .955          & .896          & .025          & .911          & .892          & .951          & .905          & .027          \\
\hline
MAGNet$_{24}$& RGB-D & 16M+335M  & .893 & .850 & .935 & .893 & .039 & .870 & .820 & .923 & .870 & .049 & .869 & .823 & .920 & .878 & .043 \\
CPNet$_{24}$& RGB-D   & 216M+335M & .922 & .887 & .950 & .916 & \underline {.028} & .895 & .856 & .935 & .896 &  {.039} & .893 & .854 & .931 & .898 &  {.035} \\

MVANet$^*$$_{24}$&   RGB-D    & 93M+335M     &  {.929}    &  {.889}    &  {.949}    &  {.919}    & .029    &  {.905}    &  {.856}    &  {.932}    & .900          &  {.040}    &  {.906}    &  {.862}    &  {.935}    &  {.907}    &  {.033}    \\

% %\rowcolor[HTML]{C9E4B4} 
\hline
\textbf{PDFNet-S} &RGB-D  &94M+24M& \underline{.932}    & .893          & \underline{.955}    & \underline{.924}    & \underline{.028}    & {.906}    & .860          & .938          & \underline{.906}    & .039          & .910          & .868          & {.941}    & .911          & {.032}    \\
\textbf{PDFNet-B} &RGB-D  &94M+97M& \textbf{.936} & \underline{.899}    & \textbf{.957} & \textbf{.928} & \textbf{.027} & \underline{.911} & \underline{.866}    & {.940}    & \textbf{.910} & \underline{.038}    & \underline{.914}    & \underline{.873}    & \underline{.943} & \underline{.914}    & \underline{.031} \\
\textbf{PDFNet-L}& RGB-D  & 94M+335M & \textbf{.936} & \textbf{.901} & \textbf{.957} & \textbf{.928} & \textbf{.027} & \textbf {.912} & \textbf{.869} & \underline {.941} & \textbf{.910} & \textbf{.037} & \textbf{.915} & \textbf{.876} & \textbf{.944} & \textbf{.916} & \textbf{.030} \\

 % \hline
 \bottomrule
\end{tabular}
\end{table*}
%-------------------------------------------------------------------------

\section{Experiments and Results}
\subsection{Data Setup and Metrics}
We conducted experiments on DIS-5K \cite{IS-net}. DIS-5K consists of 5,470 images and 225 categories in total. It contains 6 subsets, namely DIS-TR, DIS-VD, DIS-TE(1-4). Among them, DIS-TR and DIS-VD are composed of 3,000 and 470 samples respectively. DIS-TE(1-4) are ordered from simple to complex, and each TE has 500 samples.
We evaluate our method using five standard metrics, following the protocol of recent works \cite{MVANet,diffdis}: max F-measure ($F^{\text{max}}_\beta$) \cite{maxf1}, weighted F-measure ($F^{w}_\beta$) \cite{wf1}, structural measure ($S_{\alpha}$) \cite{smeasure}, E-measure ($E^m_\phi$) \cite{emeasure1,emeasure2}, and Mean Absolute Error ($M$) \cite{MAE}. $F^{\text{max}}_\beta$ and $F^{w}_\beta$ balance precision and recall, with $\beta=0.3$ prioritizing precision. $S_{\alpha}$ and $E^m_\phi$ measure structural and statistical correspondence between the predicted map and the ground truth. $M$ provides a direct measure of pixel-wise error. Higher scores are better for all metrics, except for $M$.

\subsection{Implementation Details}
All experiments were conducted on an RTX-4090. We used a Swin-B \cite{swin} backbone pre-trained on ImageNet-21K \cite{imagenet21k}. For training, all input images were resized to $1024^2$. We applied standard data augmentations, including random horizontal flipping, rotation and color jitter. The model was trained for 100 epochs using the AdamW optimizer with the learning rate of $1\times10^{-5}$ and a batch size of 1.

We generate pseudo-depth maps using three versions of DAM-v2 \cite{depth_anything_v2} (Small 24M, Base 97M, and Large 335M). The depth maps serving as model input are produced via the default $518^2 \rightarrow 1024^2$ pipeline. In contrast, a higher-fidelity map from a $1024^2 \rightarrow 1024^2$ pipeline is used as the supervision target for our depth refinement and $l_{inte}$. The S, B, and L models require 47ms, 91ms, and 127ms per inference at the default pipeline, respectively.

\subsection{Comparison with SOTAs}
\textbf{Quantitative Evaluation}. 
As shown in \tabref{tab:comparewithsota}, we compare PDFNet with SOTA models across non-diffusion and diffusion methods. 
Non-diffusion methods include IS-Net \cite{IS-net} (U$^2$-Net \cite{u2net}) and UDUN \cite{UDUN} (ResNet-50 \cite{resnet} pre-trained on ImageNet21K \cite{imagenet21k}), InSPyReNet \cite{InSPyReNet}, MVANet \cite{MVANet} (Swin-B), and BiRefNet \cite{BiRefNet} (Swin-L) (Swin pre-trained on ImageNet21K). Diffusion methods include GenPercept \cite{genpercept} (Stable Diffusion U-Net v2.1 \cite{sdunet}) and DiffDIS \cite{diffdis} (Stable Diffusion-Turbo). 
%

%%%%%%%%%%%%%%%%%%%%% Figure3:vcompare %%%%%%%%%%%%%%%%%%%%%
\begin{figure*}
  \centering
  \includegraphics[width=0.98\textwidth]{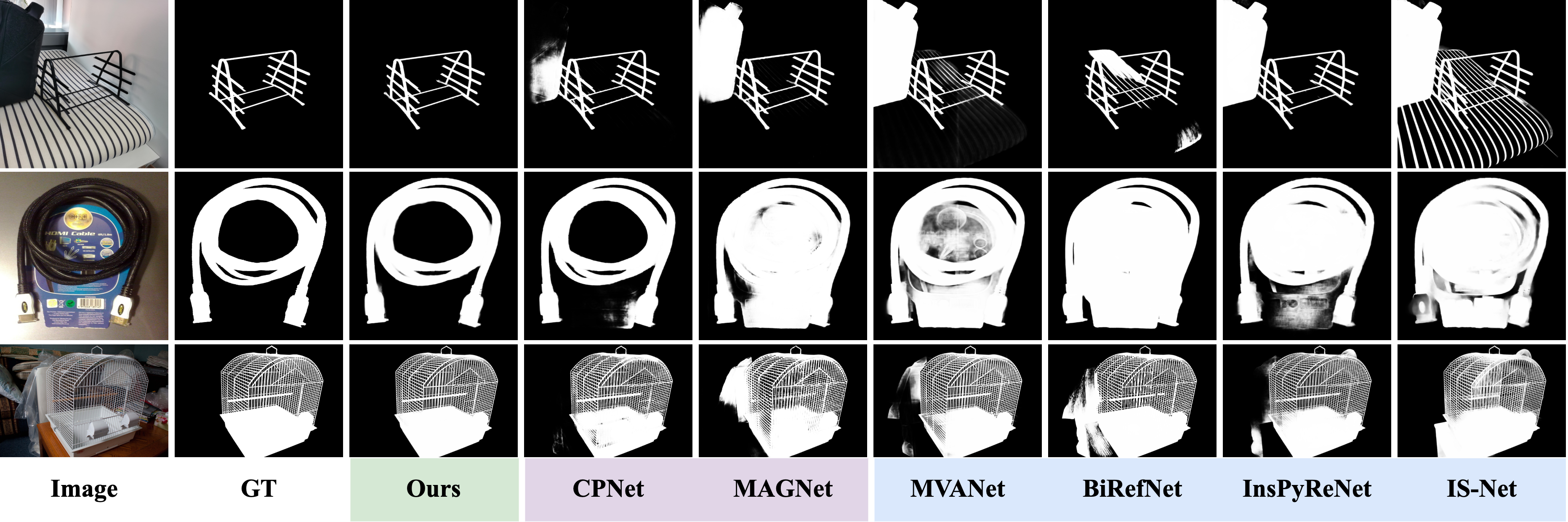}
  \caption{
Visual comparison of different DIS and RGB-D SOD methods.
 }
 % \vspace{-5pt}
  \label{fig:vcompare}
\end{figure*}

%%%%%%%%%%%%%%%%%%%%% Figure %%%%%%%%%%%%%%%%%%%%%

As no RGB-D DIS method existed before, for fairness, we evaluate other-domain SOTA RGB-D models and also adapted the best DIS model to RGB-D fashion by identical training settings. We evaluate RGB-D SOD methods MAGNet \cite{magnet}(SMT-T and MbNet-V2) and CPNet \cite{CPNet} (two Swin-B modules). Additionally, we concatenate RGB and depth as input for MVANet to create MVANet$^*$. PDFNet-S/B/L denote variants of the trained PDFNet model that, during inference, employ the DAM-v2 Small/Base/Large, respectively, as their pseudo-depth generators.

We report the number of params required at inference for all models in the table. Diffusion-based methods require both a U-Net and a VAE \cite{genpercept,diffdis}. RGB-D models additionally rely on an external depth estimator; for fair comparison, we use DAM-v2 Large \cite{depth_anything_v2}. All models are tested with $1024^2$ input as IS-Net \cite{IS-net} settings. All models use their official params and re-ran on an RTX 4090 for fairness, so they may differ from their original results.

PDFNet outperforms all non-diffusion models, surpassing MVANet by 0.7\%, 1.5\%, 0.6\%, 1.2\%, and 0.5\% on $F^{max}_\beta$, $F^{w}_\beta$, $S_{\alpha}$, $E^m_\phi$, and $M$, respectively, on DIS-TE (ALL). Notably, PDFNet also outperforms diffusion methods on several metrics while using less than 50\% of their params. FPS comparison with existing methods are as follows: BiRefNet (6), MVANet (6.5), PDFNet-S, PDFNet-B and PDFNet-L (5.7, 4.5 and 3.9, including the depth generation time), and DiffDIS (0.8). 

\noindent\textbf{Qualitative Evaluation}.
As shown in \figref{fig:vcompare}, our depth integrity-prior provides crucial structural guidance. For instance (row 1), while other methods mistakenly segment the sofa's pattern, ours correctly delineates the foreground shelf, demonstrating an understanding of scene structure. 
%-----------------------------------ablation:2 modules--------------------------------------
\begin{table}
\caption{Ablation experiments of components.}
\label{tab:ablation-modules}
\vspace{-0.2cm}
\centering
\small
\renewcommand{\arraystretch}{1}
\renewcommand{\tabcolsep}{0.75mm}
\begin{tabular}{cccc|ccccc|c}
\toprule
$\textbf{S}$& $\textbf{Bd}$& \textbf{FSE} & \textbf{Depth}  & $F^{max}_\beta$ & $F^w_\beta$   & $E_{\phi}^{m}$ & $S_{\alpha}$ & $M$ & $FPS$   \\ \hline 
  & & &   & .841          &.779 & .884         & .842          & .057          & \textbf{7.300} \\
 & &  & \checkmark & .872          &.832 & .916    & .877    &  .044    & 4.525          \\
 & & \checkmark &   & .885    &.847 & .920    &  .882    & .043    & \underline{ 6.273}    \\
\checkmark& &\checkmark&  &  .890    &.850 &  .921    &  .890    &  .041    & 6.021\\
   &\checkmark&\checkmark&  & .891   &.852 & .922    &  .891    &   .038    &  6.113    \\
   \checkmark & \checkmark & \checkmark & & \underline{.903} & \underline{ .857}    & \underline{ .937}    & \underline{ .907}    & \underline{ .036}    & 6.043    \\
% %\rowcolor[HTML]{C9E4B4} 
\checkmark &\checkmark & \checkmark & \checkmark & \textbf{.907}          &\textbf{.866} & \textbf{.940}          & \textbf{.912}    & \textbf{.032} & 3.925          \\
% \hline
\bottomrule
\end{tabular}
\end{table}
%-------------------------------------------------------------------------

%-----------------------------------ablation:2 modules--------------------------------------
\begin{table}[]
\caption{Ablation experiments of depth loss.}
\label{tab:ablation-loss}
\vspace{-0.2cm}
\centering\small
\renewcommand{\arraystretch}{1}
\renewcommand{\tabcolsep}{2.3mm}
\begin{tabular}{cc|ccccc}
\toprule
$L_{SILog}$ & $L_{inte}$  & $F^{max}_\beta$ &$F^w_\beta$  & $E_{\phi}^{m}$ & $S_{\alpha}$ & $M$    \\ 
\hline 
&   & .907  & .866        & .940          & .912    & {.032}    \\
\checkmark &   & .909    & .867      & .939          & \underline{.913}          & {.032}          \\
& \checkmark & \underline{.912}  & \underline{.869}  & \underline{.943}    & \textbf{.916} & \underline{.030} \\
%\rowcolor[HTML]{C9E4B4} 
\checkmark & \checkmark & \textbf{.915} & \textbf{.875} & \textbf{.945} & \textbf{.916} & \textbf{.029}
\\
% \hline
\bottomrule
\end{tabular}
\end{table}
%-------------------------------------------------------------------------

%%%%%%%%%%%%%%%%%%%%% Figure2:overall %%%%%%%%%%%%%%%%%%%%%
\begin{figure}[]
  \centering
  \includegraphics[width=0.46\textwidth]{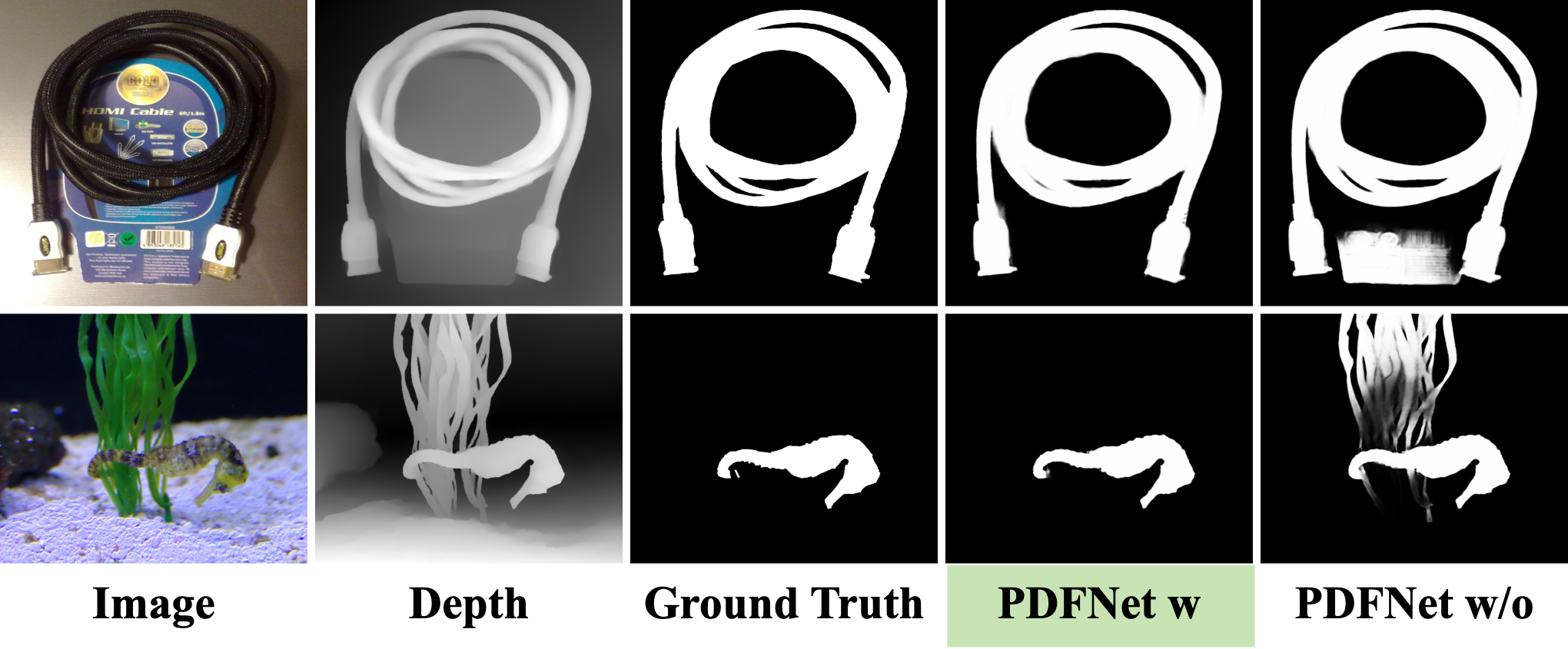}
  
  \caption{Comparison before and after adding $l_{inte}$. PDFNet w (PDFNet with $l_{inte}$) vs. PDFNet w/o (PDFNet without $l_{inte}$).
 }
\vspace{-0.2cm}
  \label{fig:ablation-linte}
\end{figure}

%%%%%%%%%%%%%%%%%%%%% Figure %%%%%%%%%%%%%%%%%%%%%

\subsection{Ablation Study}
In this section, we analyze the influence of each module on the performance and the impact of pseudo depth map quality. All experiments were conducted only on DIS-VD. 

\subsubsection{Components of PDFNet}
\textbf{Baseline}.
We use a simple encoder-decoder baseline with 8$\times$8 patches. In the models without $S$ and $Bd$, we replace their computations with all-ones matrices, respectively. For the model without FSE, encoder outputs are passed through a Conv3$\times$3, SiLU, and RMSNorm layer before entering the decoder. For the model without depth, we set the depth input to zero and exclude depth generation time. $L_{SILog}$ and $L_{inte}$ were not used in the component ablation.

\noindent\textbf{FSE.} 
As shown in \tabref{tab:ablation-modules}, the FSE module improves modality fusion and sharpens details through patch selection and feature fusion, leading to get better $F^{max}_\beta$.

\noindent\textbf{Depth.} 
Depth provides integrity-prior throughout the network, significantly enhancing target integrity detection.

\subsubsection{Depth Loss}
As shown in \tabref{tab:ablation-loss}, each component improves performance. Combining depth refinement with the $l_{inte}$ brings more gain than using it alone, indicating that the $l_{inte}$ helps the model focus on smooth and consistent depth regions during mask prediction.
We selected samples to compare the results before and after using $l_{inte}$ as shown in \figref{fig:ablation-linte}. It can be observed that after adding $l_{inte}$, the model has a better grasp of the depth integrity.
To investigate the generalizability of the $l_{inte}$, we conducted experiments on different models such as MAGNet \cite{magnet}, CPNet \cite{CPNet}, and PDFNet. As shown in \tabref{tab:IPLoss-abc}, $l_{inte}$ can improve on other models.

%-----------------------------------ablation:2 L_inte on other models--------------------------------------
\begin{table}
\caption{Whether to employ $L_{inte}$ for supervision.}
\label{tab:IPLoss-abc}
\vspace{-0.2cm}
\centering
\small
\renewcommand{\arraystretch}{1}
\renewcommand{\tabcolsep}{2.5mm}
\begin{tabular}{c|ccccc}
\toprule
\textbf{Method} &$F^{max}_\beta$ &$F^w_\beta$ &$E_{\phi}^{m}$ &$S_{\alpha}$ &$~M~$\\
\hline

MAGNet W/O  & .867          & .820          & .917          & .879          & .045          \\
MAGNet W  & .870          & .825          & .920          & .881          & .043          \\
\hline
CPNet W/O & .892 & .855 & .933 & .900 & .034 \\
CPNet W & .895 & .857 & .935 & .902 & .033 \\
\hline

{PDFNet W/O}  & \underline{.907}          & \underline{.866} & \underline{.940} & \underline{.912} & \underline{.032} \\
% \rowcolor[HTML]{C9E4B4} 
\textbf{PDFNet W}  & \textbf{.912}          & \textbf{.869} & \textbf{.943} & \textbf{.916} & \textbf{.030} \\
\bottomrule
\end{tabular}
% \vspace{-5pt}
\end{table}
%-------------------------------------------------------------------------

\subsubsection{Number of Patches} 
To analyze the effect of patch density, we partition the image into 1, 4, 16, 64, and 256 patches. As shown in \tabref{tab:ablation-patch}, our performance peaks at 64 patches, unlike prior work. This is because our architecture preserves global context through a full-resolution main branch, while the patch branch enhances local detail by restricting the receptive field. At 256 patches, performance declines, likely because the receptive field becomes overly constrained, losing the context needed to capture fine edge details. FPS measurements in this experiment exclude depth generation time.

%-----------------------------------ablation:3 patches nums--------------------------------------
\begin{table}[]
\caption{Ablation of number of patches.}
\label{tab:ablation-patch}
\vspace{-0.2cm}
\centering
\small
\renewcommand{\arraystretch}{1}
\renewcommand{\tabcolsep}{1.3mm}
\begin{tabular}{c|ccccc|c}
\toprule
\textbf{Patches Number} & $F^{max}_\beta$ & $F^w_\beta$ & $E_{\phi}^{m}$ & $S_{\alpha}$ & $M$ & $FPS$   \\ 
\hline
MVANet 2$\times$2   & .904     & .856     & .938          & .905          & .036          & 6.534 \\
MVANet 3$\times$3   & .803     & .742     & .824          & .782          & .058          & 6.427 \\
MVANet 4$\times$4   & .707     & .634     & .749          & .722          & .088          & 6.362 \\
\hline
PDFNet 1$\times$1   & .907     & .866     & .939          & .909          & .032          & \textbf{7.230} \\
PDFNet 2$\times$2   & .908     & .866     & .941          & .911          & \underline{ .031}    & \underline{ 7.153}    \\
PDFNet 4$\times$4   & \underline{ .911}  & \underline{ .868}  & \underline{ .943}    & \underline{ .913}    & \underline{ .031}    & 6.502          \\
%\rowcolor[HTML]{C9E4B4} 
\textbf{PDFNet 8$\times$8}   & \textbf{.915} & \textbf{.875}  & \textbf{.945} & \textbf{.916} & \textbf{.029} & 6.043          \\
PDFNet 16$\times$16 & .910     & .867     & .941          & .912          & \underline{ .031}    & 3.375         
\\ 
\bottomrule
\end{tabular}
\end{table}
%-------------------------------------------------------------------------

%-----------------------------------HRSOD Task--------------------------------------
\begin{table}[]
\caption{Comparisons of PDFNet with PGNet \cite{pyramid}, InSPyReNet \cite{InSPyReNet} and BiRefNet \cite{BiRefNet} on HRSOD \cite{HRSOD} and UHRSO \cite{pyramid}.}
\vspace{-0.2cm}
\centering
\small
\renewcommand{\arraystretch}{1}
\renewcommand{\tabcolsep}{0.1mm}
\begin{tabular}{c|ccccc|ccccc}
 \toprule
\multirow{2}{*}{\textbf{Method}}  & \multicolumn{5}{c}{\textbf{HRSOD-TE (400)}} & \multicolumn{5}{|c}{\textbf{UHRSD-TE (988)}}\\
~&$F^{max}_\beta$&$F^w_\beta$&$E_{\phi}^{m}$&$S_{\alpha}$&$M$&$F^{max}_\beta$&$F^w_\beta$&$E_{\phi}^{m}$&$S_{\alpha}$&$M$ \\
 \hline
PGNet$_{22}$               & .939          & .901          & .960          & .938          & .020          & .939          & .917          & .955          & .935          & .026          \\
InSPyReNet$_{22}$        & \underline{.955}    & .922          & .964          & \underline{.956}    & .018          & .957          & .934          & .962          & \textbf{.953} & \underline{.020}     \\
BiRefNet$_{24}$           & .952          & \underline{.931}    & \underline{.967}    & \underline{.956}    & \underline{.016}    & \underline{.958}    & \underline{.941}    & \underline{.964}    & \underline{.952}    & \textbf{.019} \\
% \rowcolor[HTML]{C9E4B4} 
\hline
\textbf{PDFNet-L}    & \textbf{.965} & \textbf{.943} & \textbf{.977} & \textbf{.963} & \textbf{.012} & \textbf{.963} & \textbf{.945} & \textbf{.966} & \textbf{.953} & \textbf{.019}
\\
 \bottomrule
\end{tabular}
% \vspace{-5pt}
\label{tab:SOD-tab}
\end{table}
%-------------------------------------------------------------------------

\subsection{Generalizability and Robustness}
\noindent\textbf{Validation on HRSOD.} We conducted additional experiments on the High-Resolution Salient Object Detection (HRSOD) task. Following~\cite{BiRefNet,pyramid}, PDFNet was trained on a combined set of HRSOD-TR \cite{HRSOD} (1,610 samples) and UHRSD-TR \cite{pyramid} (4,936 samples), with implementation details identical to those of the DIS task except for reducing the training epochs to 40. We evaluated our method on HRSOD-TE (400 samples) and UHRSD-TE (988 samples), comparing against PGNet \cite{pyramid}, InSPyReNet \cite{InSPyReNet}, and BiRefNet \cite{BiRefNet}, as shown in \tabref{tab:SOD-tab}. With the same training data, PDFNet surpasses previous approaches on these benchmarks. 

These results show the remarkable generalization ability of our PDFNet to similar HR tasks.

\noindent\textbf{Validation on different depth generators.} We employ different pseudo-depth generators to produce pseudo-depth inputs for the trained PDFNet, and the results are shown in \tabref{tab:diff-depth-generator}. The pseudo-depth generators include DAM-Small \cite{depth_anything_v1}, DAM-v2-Small/Base/Large \cite{depth_anything_v2} and DepthPro \cite{depthpro}. These results show that the trained PDFNet demonstrates high robustness to pseudo-depth inputs of varying quality.

%-----------------------------------diff depth generator--------------------------------------
\begin{table}[]
\caption{Performance of the trained PDFNet using different depth generated by different generators on DIS-VD.}
\vspace{-0.2cm}
\centering
\small
\renewcommand{\arraystretch}{1}
\renewcommand{\tabcolsep}{1.45mm}
\begin{tabular}{c|c|ccccc}
\toprule
\textbf{Depth Generator}& \textbf{Params} &$F^{max}_\beta$ &$F^w_\beta$ &$E_{\phi}^{m}$ &$S_{\alpha}$ &$~M~$\\
\hline
DAM S & 24M & .904 & .860 & .938& .906 & .033 \\
\hline
DAM-v2 S & 24M & .909& .868 & .942& \underline{.913} & \underline{.030} \\
DAM-v2 B & 97M & .912& .873 & \underline{.944}& \textbf{.916} & \underline{.030} \\
DAM-v2 L & 335M & \underline{.915}& \underline{.875} & \textbf{.945}& \textbf{.916} & \textbf{.029} \\
\hline
Depthpro & 1B & \textbf{.917} & \textbf{.876} & \textbf{.945} & \textbf{.916} & \underline{.030} \\
\bottomrule
\end{tabular}
% \vspace{-10pt}
\label{tab:diff-depth-generator}
\end{table}
%-------------------------------------------------------------------------

\section{Conclusion}
This paper introduces depth information into the DIS field to guide object structure understanding and proposes PDFNet, which leverages pseudo-depth information to perform both DIS and HRSOD tasks within a unified framework. Through experiments, we observe a significant variance difference between the GT regions and background regions in depth images—a property we term the depth integrity-prior. To exploit this prior, we enhance the consistency between segmentation results and depth via a depth integrity-prior loss. Additionally, we fuse depth features with patch features using the FSE module and sharpen object boundaries. Extensive experiments demonstrate that our PDFNet achieves strong performance on DIS and other similar high-resolution tasks, fully validating the outstanding performance and strong generalization capability of PDFNet, and offering a valuable solution to the academic community. 
Given the prevalence of multimodal learning, this offers a reference for future DIS research.

{
    \small
    \bibliographystyle{ieeenat_fullname}
    \bibliography{main}

@String(CVPR= {IEEE Conf. Comput. Vis. Pattern Recog.})

@String(ICCV= {Int. Conf. Comput. Vis.})

@String(IJCAI = {IJCAI})

@String(AAAI = {AAAI})

@String(VR   = {Vis. Res.})

@String(CVPR  = {CVPR})

@String(ICCV  = {ICCV})

@inproceedings{IS-net,
  title={Highly accurate dichotomous image segmentation},
  author={Qin, Xuebin and Dai, Hang and Hu, Xiaobin and Fan, Deng-Ping and Shao, Ling and Van Gool, Luc},
  booktitle={European Conference on Computer Vision},
  pages={38--56},
  year={2022},
  organization={Springer}
}

@inproceedings{FP-DIS,
  title={Dichotomous Image Segmentation with Frequency Priors.},
  author={Zhou, Yan and Dong, Bo and Wu, Yuanfeng and Zhu, Wentao and Chen, Geng and Zhang, Yanning},
  booktitle={IJCAI},
  volume={1},
  number={2},
  pages={3},
  year={2023}
}

@inproceedings{UDUN,
  title={Unite-divide-unite: Joint boosting trunk and structure for high-accuracy dichotomous image segmentation},
  author={Pei, Jialun and Zhou, Zhangjun and Jin, Yueming and Tang, He and Heng, Pheng-Ann},
  booktitle={Proceedings of the 31st ACM International Conference on Multimedia},
  pages={2139--2147},
  year={2023}
}

@inproceedings{MVANet,
  title={Multi-view Aggregation Network for Dichotomous Image Segmentation},
  author={Yu, Qian and Zhao, Xiaoqi and Pang, Youwei and Zhang, Lihe and Lu, Huchuan},
  booktitle={Proceedings of the IEEE/CVF Conference on Computer Vision and Pattern Recognition},
  pages={3921--3930},
  year={2024}
}

@article{BiRefNet,
  title={Bilateral Reference for High-Resolution Dichotomous Image Segmentation},
  author={Zheng, Peng and Gao, Dehong and Fan, Deng-Ping and Liu, Li and Laaksonen, Jorma and Ouyang, Wanli and Sebe, Nicu},
  journal={CAAI Artificial Intelligence Research},
  volume = {3},
  pages = {9150038},
  year={2024}
}

@inproceedings{
depth_anything_v2,
title={Depth Anything V2},
author={Lihe Yang and Bingyi Kang and Zilong Huang and Zhen Zhao and Xiaogang Xu and Jiashi Feng and Hengshuang Zhao},
booktitle={The Thirty-eighth Annual Conference on Neural Information Processing Systems},
year={2024},
url={https://openreview.net/forum?id=cFTi3gLJ1X}
}

@inproceedings{depth_anything_v1,
  title={Depth Anything: Unleashing the Power of Large-Scale Unlabeled Data}, 
  author={Yang, Lihe and Kang, Bingyi and Huang, Zilong and Xu, Xiaogang and Feng, Jiashi and Zhao, Hengshuang},
  booktitle={CVPR},
  year={2024}
}

@inproceedings{monodepth17,
  title     = {Unsupervised Monocular Depth Estimation with Left-Right Consistency},
  author    = {Cl{\'{e}}ment Godard and
               Oisin {Mac Aodha} and
               Gabriel J. Brostow},
  booktitle = {CVPR},
  year = {2017}
}

@inproceedings{impact,
  title={Impact of pseudo depth on open world object segmentation with minimal user guidance},
  author={Sch{\"o}n, Robin and Ludwig, Katja and Lienhart, Rainer},
  booktitle={Proceedings of the IEEE/CVF Conference on Computer Vision and Pattern Recognition},
  pages={4809--4819},
  year={2023}
}

@article{polyp-DAM,
  title={Polyp-DAM: Polyp segmentation via depth anything model},
  author={Zheng, Zhuoran and Wu, Chen and Jin, Yeying and Jia, Xiuyi},
  journal={IEEE Signal Processing Letters},
  year={2024},
  publisher={IEEE}
}

@inproceedings{IOCUS,
    title={Indiscernible Object Counting in Underwater Scenes},
    author={Sun, Guolei and An, Zhaochong and Liu, Yun and Liu, Ce and Sakaridis, Christos and Fan, Deng-Ping and Van Gool, Luc},
    booktitle={Proceedings of the IEEE/CVF International Conference on Computer Vision and Patern Recognition (CVPR)},
    year={2023}
}

@article{SwiGLu,
  title={Glu variants improve transformer},
  author={Shazeer, Noam},
  journal={arXiv preprint arXiv:2002.05202},
  year={2020}
}

@article{RMSNorm,
  title={Root mean square layer normalization},
  author={Zhang, Biao and Sennrich, Rico},
  journal={Advances in Neural Information Processing Systems},
  volume={32},
  year={2019}
}

@inproceedings{maxf1,
  title={Frequency-tuned salient region detection},
  author={Achanta, Radhakrishna and Hemami, Sheila and Estrada, Francisco and Susstrunk, Sabine},
  booktitle={2009 IEEE conference on computer vision and pattern recognition},
  pages={1597--1604},
  year={2009},
  organization={IEEE}
}

@inproceedings{wf1,
  title={How to evaluate foreground maps?},
  author={Margolin, Ran and Zelnik-Manor, Lihi and Tal, Ayellet},
  booktitle={Proceedings of the IEEE conference on computer vision and pattern recognition},
  pages={248--255},
  year={2014}
}

@inproceedings{MAE,
  title={Saliency filters: Contrast based filtering for salient region detection},
  author={Perazzi, Federico and Kr{\"a}henb{\"u}hl, Philipp and Pritch, Yael and Hornung, Alexander},
  booktitle={2012 IEEE conference on computer vision and pattern recognition},
  pages={733--740},
  year={2012},
  organization={IEEE}
}

@inproceedings{smeasure,
  title={Structure-measure: A new way to evaluate foreground maps},
  author={Fan, Deng-Ping and Cheng, Ming-Ming and Liu, Yun and Li, Tao and Borji, Ali},
  booktitle={Proceedings of the IEEE international conference on computer vision},
  pages={4548--4557},
  year={2017}
}

@inproceedings{emeasure1,
  title={Enhanced-alignment measure for binary foreground map evaluation},
  author={Fan, Deng-Ping and Gong, Cheng and Cao, Yang and Ren, Bo and Cheng, Ming-Ming and Borji, Ali},
  booktitle={Proceedings of the 27th International Joint Conference on Artificial Intelligence},
  pages={698--704},
  year={2018}
}

@article{emeasure2,
  title={Cognitive vision inspired object segmentation metric and loss function},
  author={Fan, Deng-Ping and Ji, Ge-Peng and Qin, Xuebin and Cheng, Ming-Ming},
  journal={Scientia Sinica Informationis},
  volume={6},
  number={6},
  year={2021}
}

@inproceedings{swin,
  title={Swin Transformer: Hierarchical Vision Transformer using Shifted Windows},
  author={Liu, Ze and Lin, Yutong and Cao, Yue and Hu, Han and Wei, Yixuan and Zhang, Zheng and Lin, Stephen and Guo, Baining},
  booktitle={Proceedings of the IEEE/CVF International Conference on Computer Vision (ICCV)},
  year={2021}
}

@inproceedings{wbce-wiou,
  title={F$^3$Net: fusion, feedback and focus for salient object detection},
  author={Wei, Jun and Wang, Shuhui and Huang, Qingming},
  booktitle={Proceedings of the AAAI conference on artificial intelligence},
  volume={34},
  number={07},
  pages={12321--12328},
  year={2020}
}

@article{ssim,
  title={Image quality assessment: from error visibility to structural similarity},
  author={Wang, Zhou and Bovik, Alan C and Sheikh, Hamid R and Simoncelli, Eero P},
  journal={IEEE transactions on image processing},
  volume={13},
  number={4},
  pages={600--612},
  year={2004},
  publisher={IEEE}
}

@inproceedings{genpercept,
title={What Matters When Repurposing Diffusion Models for General Dense Perception Tasks?},
author={Guangkai Xu and Yongtao Ge and Mingyu Liu and Chengxiang Fan and Kangyang Xie and Zhiyue Zhao and Hao Chen and Chunhua Shen},
booktitle={The Thirteenth International Conference on Learning Representations},
year={2025},
url={https://openreview.net/forum?id=BgYbk6ZmeX}
}

@inproceedings{diffdis,
title={High-Precision Dichotomous Image Segmentation via Probing Diffusion Capacity},
author={Qian Yu and Peng-Tao Jiang and Hao Zhang and Jinwei Chen and Bo Li and Lihe Zhang and Huchuan Lu},
booktitle={The Thirteenth International Conference on Learning Representations},
year={2025},
url={https://openreview.net/forum?id=vh1e2WJfZp}
}

@inproceedings{InSPyReNet,
  title={Revisiting image pyramid structure for high resolution salient object detection},
  author={Kim, Taehun and Kim, Kunhee and Lee, Joonyeong and Cha, Dongmin and Lee, Jiho and Kim, Daijin},
  booktitle={Proceedings of the Asian Conference on Computer Vision},
  pages={108--124},
  year={2022}
}

@inproceedings{
depthpro,
title={Depth Pro: Sharp Monocular Metric Depth in Less Than a Second},
author={Alexey Bochkovskiy and Ama{\"e}l Delaunoy and Hugo Germain and Marcel Santos and Yichao Zhou and Stephan Richter and Vladlen Koltun},
booktitle={The Thirteenth International Conference on Learning Representations},
year={2025},
url={https://openreview.net/forum?id=aueXfY0Clv}
}

@article{magnet,
  title={MAGNet: Multi-scale Awareness and Global fusion Network for RGB-D salient object detection},
  author={Zhong, Mingyu and Sun, Jing and Ren, Peng and Wang, Fasheng and Sun, Fuming},
  journal={Knowledge-Based Systems},
  pages={112126},
  year={2024},
  publisher={Elsevier}
}

@article{CPNet,
  title={Cross-modal fusion and progressive decoding network for RGB-D salient object detection},
  author={Hu, Xihang and Sun, Fuming and Sun, Jing and Wang, Fasheng and Li, Haojie},
  journal={International Journal of Computer Vision},
  pages={1--19},
  year={2024},
  publisher={Springer}
}

@article{u2net,
  title={U2-Net: Going deeper with nested U-structure for salient object detection},
  author={Qin, Xuebin and Zhang, Zichen and Huang, Chenyang and Dehghan, Masood and Zaiane, Osmar R and Jagersand, Martin},
  journal={Pattern recognition},
  volume={106},
  pages={107404},
  year={2020},
  publisher={Elsevier}
}

@article{SiLU,
  title={Sigmoid-weighted linear units for neural network function approximation in reinforcement learning},
  author={Elfwing, Stefan and Uchibe, Eiji and Doya, Kenji},
  journal={Neural networks},
  volume={107},
  pages={3--11},
  year={2018},
  publisher={Elsevier}
}

@inproceedings{swinir,
  title={Swinir: Image restoration using swin transformer},
  author={Liang, Jingyun and Cao, Jiezhang and Sun, Guolei and Zhang, Kai and Van Gool, Luc and Timofte, Radu},
  booktitle={Proceedings of the IEEE/CVF international conference on computer vision},
  pages={1833--1844},
  year={2021}
}

@article{p2t,
  title={P2T: Pyramid pooling transformer for scene understanding},
  author={Wu, Yu-Huan and Liu, Yun and Zhan, Xin and Cheng, Ming-Ming},
  journal={IEEE transactions on pattern analysis and machine intelligence},
  volume={45},
  number={11},
  pages={12760--12771},
  year={2022},
  publisher={IEEE}
}

@inproceedings{asymmetric,
  title={Asymmetric non-local neural networks for semantic segmentation},
  author={Zhu, Zhen and Xu, Mengde and Bai, Song and Huang, Tengteng and Bai, Xiang},
  booktitle={Proceedings of the IEEE/CVF international conference on computer vision},
  pages={593--602},
  year={2019}
}

@article{SILogloss,
  title={Towards robust monocular depth estimation: Mixing datasets for zero-shot cross-dataset transfer},
  author={Ranftl, Ren{\'e} and Lasinger, Katrin and Hafner, David and Schindler, Konrad and Koltun, Vladlen},
  journal={IEEE transactions on pattern analysis and machine intelligence},
  volume={44},
  number={3},
  pages={1623--1637},
  year={2020},
  publisher={IEEE}
}

@article{boosting,
  title={Boosting box-supervised instance segmentation with pseudo depth},
  author={Yu, Xinyi and Yan, Ling and Jiang, Pengtao and Chen, Hao and Li, Bo and Wu, Lin Yuanbo and Ou, Linlin},
  journal={arXiv preprint arXiv:2403.01214},
  year={2024}
}

@inproceedings{pyramid,
  title={Pyramid grafting network for one-stage high resolution saliency detection},
  author={Xie, Chenxi and Xia, Changqun and Ma, Mingcan and Zhao, Zhirui and Chen, Xiaowu and Li, Jia},
  booktitle={Proceedings of the IEEE/CVF conference on computer vision and pattern recognition},
  pages={11717--11726},
  year={2022}
}

@InProceedings{HRSOD,
author = {Zeng, Yi and Zhang, Pingping and Zhang, Jianming and Lin, Zhe and Lu, Huchuan},
title = {Towards High-Resolution Salient Object Detection},
booktitle = {IEEE International Conference on Computer Vision (ICCV)},
month = {October},
year = {2019}
}

@inproceedings{resnet,
  title={Deep residual learning for image recognition},
  author={He, Kaiming and Zhang, Xiangyu and Ren, Shaoqing and Sun, Jian},
  booktitle={Proceedings of the IEEE conference on computer vision and pattern recognition},
  pages={770--778},
  year={2016}
}

@inproceedings{
imagenet21k,
title={ImageNet-21K Pretraining for the Masses},
author={Tal Ridnik and Emanuel Ben-Baruch and Asaf Noy and Lihi Zelnik-Manor},
booktitle={Thirty-fifth Conference on Neural Information Processing Systems Datasets and Benchmarks Track (Round 1)},
year={2021},
url={https://openreview.net/forum?id=Zkj_VcZ6ol}
}

@InProceedings{sdunet,
    author    = {Rombach, Robin and Blattmann, Andreas and Lorenz, Dominik and Esser, Patrick and Ommer, Bj\"orn},
    title     = {High-Resolution Image Synthesis With Latent Diffusion Models},
    booktitle = {Proceedings of the IEEE/CVF Conference on Computer Vision and Pattern Recognition (CVPR)},
    month     = {June},
    year      = {2022},
    pages     = {10684-10695}
}

@article{dis-imageediting,
  title={Context-aware saliency detection},
  author={Goferman, Stas and Zelnik-Manor, Lihi and Tal, Ayellet},
  journal={IEEE transactions on pattern analysis and machine intelligence},
  volume={34},
  number={10},
  pages={1915--1926},
  year={2011},
  publisher={Ieee}
}

@article{AR1,
  title={Kine-Appendage: Enhancing Freehand VR Interaction Through Transformations of Virtual Appendages},
  author={Tian, Yang and Bai, Hualong and Zhao, Shengdong and Fu, Chi-Wing and Yu, Chun and Qin, Haozhao and Wang, Qiong and Heng, Pheng-Ann},
  journal={IEEE Transactions on Visualization and Computer Graphics},
  year={2022},
  publisher={IEEE}
}

@article{AR2,
  title={Boundary-aware segmentation network for mobile and web applications},
  author={Qin, Xuebin and Fan, Deng-Ping and Huang, Chenyang and Diagne, Cyril and Zhang, Zichen and Sant'Anna, Adri{\`a} Cabeza and Suarez, Albert and Jagersand, Martin and Shao, Ling},
  journal={arXiv preprint arXiv:2101.04704},
  year={2021}
}

@inproceedings{
    IC-Light,
    title={Scaling In-the-Wild Training for Diffusion-based Illumination Harmonization and Editing by Imposing Consistent Light Transport},
    author={Lvmin Zhang and Anyi Rao and Maneesh Agrawala},
    booktitle={The Thirteenth International Conference on Learning Representations},
    year={2025},
    url={https://openreview.net/forum?id=u1cQYxRI1H}
}

@InProceedings{Segment_any_motion,
    author    = {Huang, Nan and Zheng, Wenzhao and Xu, Chenfeng and Keutzer, Kurt and Zhang, Shanghang and Kanazawa, Angjoo and Wang, Qianqian},
    title     = {Segment Any Motion in Videos},
    booktitle = {Proceedings of the Computer Vision and Pattern Recognition Conference (CVPR)},
    month     = {June},
    year      = {2025},
    pages     = {3406-3416}
}

@INPROCEEDINGS{popnet,
  title={Source-free depth for object pop-out},
  author={Wu, Zongwei and Paudel, Danda Pani and Fan, Deng-Ping and Wang, Jingjing and Wang, Shuo and Demonceaux, Cédric and Timofte, Radu and Van Gool, Luc},
  booktitle={ICCV}, 
  year={2023},
}

@INPROCEEDINGS{ssfSOD,
  author={Zhang, Miao and Ren, Weisong and Piao, Yongri and Rong, Zhengkun and Lu, Huchuan},
  booktitle={2020 IEEE/CVF Conference on Computer Vision and Pattern Recognition (CVPR)}, 
  title={Select, Supplement and Focus for RGB-D Saliency Detection}, 
  year={2020},
  volume={},
  number={},
  pages={3469-3478},
  doi={10.1109/CVPR42600.2020.00353}}
}

\end{document}